\documentclass[10pt,twocolumn,letterpaper]{article}
\usepackage[accsupp]{axessibility}
\usepackage{iccv}
\usepackage{times}
\usepackage{epsfig}
\usepackage{graphicx}
\usepackage{amsmath}
\usepackage{amssymb}

\usepackage{multirow}
\usepackage{mwe}
\usepackage{subcaption,float}
\usepackage{booktabs}

\usepackage[pagebackref=true,breaklinks=true,letterpaper=true,colorlinks,bookmarks=false]{hyperref}
\iccvfinalcopy 

\def\iccvPaperID{4215} 

\ificcvfinal\fi

\begin{document}

\title{MMVP: Motion-Matrix-based Video Prediction}

\author{Yiqi Zhong$^1$\footnotemark[1]  \footnotemark[2]\quad 
Luming Liang$^2$\footnotemark[1] 
 \footnotemark[3]\quad 
Ilya Zharkov$^2$\ \quad 
Ulrich Neumann$^1$\footnotemark[3]\\
$^{1}${University of Southern California} \quad $^{2}${Microsoft} \\ 
$^{1}${\tt\small \{yiqizhon, uneumann\}@usc.edu}  \quad 
$^{2}${\tt\small \{lulian,zharkov\}@microsoft.com} \\
}

\maketitle
\ificcvfinal\thispagestyle{empty}\fi

\renewcommand{\thefootnote}{\fnsymbol{footnote}}
\footnotetext[1]{Equal contributions.} 
\footnotetext[2]{The work is done during Yiqi Zhong's internship at Microsoft.}
\footnotetext[3]
{Corresponding authors.}

\begin{abstract}
   A central challenge of video prediction lies where the system has to reason the objects' future motions from image frames while simultaneously maintaining the consistency of their appearances across frames. This work introduces an end-to-end trainable two-stream video prediction framework, Motion-Matrix-based Video Prediction (MMVP), to tackle this challenge. Unlike previous methods that usually handle motion prediction and appearance maintenance within the same set of modules, MMVP decouples motion and appearance information by constructing appearance-agnostic motion matrices. The motion matrices represent the temporal similarity of each and every pair of feature patches in the input frames, and are the sole input of the motion prediction module in MMVP. This design improves video prediction in both accuracy and efficiency, and reduces the model size. Results of extensive experiments demonstrate that MMVP outperforms state-of-the-art systems on public data sets by non-negligible large margins ($\approx$ 1 db in PSNR, UCF Sports) in significantly smaller model sizes ($84\%$ the size or smaller). Please refer to this \href{https://github.com/Kay1794/MMVP-motion-matrix-based-video-prediction}{link} for the official code and the datasets used in this paper.

   
\end{abstract}

\section{Introduction}
\label{sec:intro}

Video prediction aims at predicting future frames from limited past frames. It is a longstanding yet unsolved problem studied for decades~\cite{girod1987efficiency,chang1997optimal}. Advancing research in this area benefits various applications such as video compression~\cite{liu2021deep,yang2022advancing,lin2020m}, surveillance systems~\cite{zhang2022surveillance,zhang2013background,duque2007prediction}, and robotics~\cite{finn2017deep,hirose2019deep,ebert2018visual}. The task can be essentially broken down into two sub-tasks: i) motion prediction and ii) frame synthesis. Each sub-task has its unique goal that cannot be simply accomplished by achieving the other one. For the sub-task of \emph{motion prediction}, systems need to reason the future movements of objects/backgrounds by discovering the motion cues hidden in the past frames. Whereas for the sub-task of \emph{frame synthesis}, systems need to maintain appearance features and generate future frames that keep appearance consistency. These separated goals make video prediction inherently much more difficult than the individual task of normal motion prediction or content synthesis.

\begin{figure}[h!]
    \centering
    \includegraphics[width=0.4\textwidth]{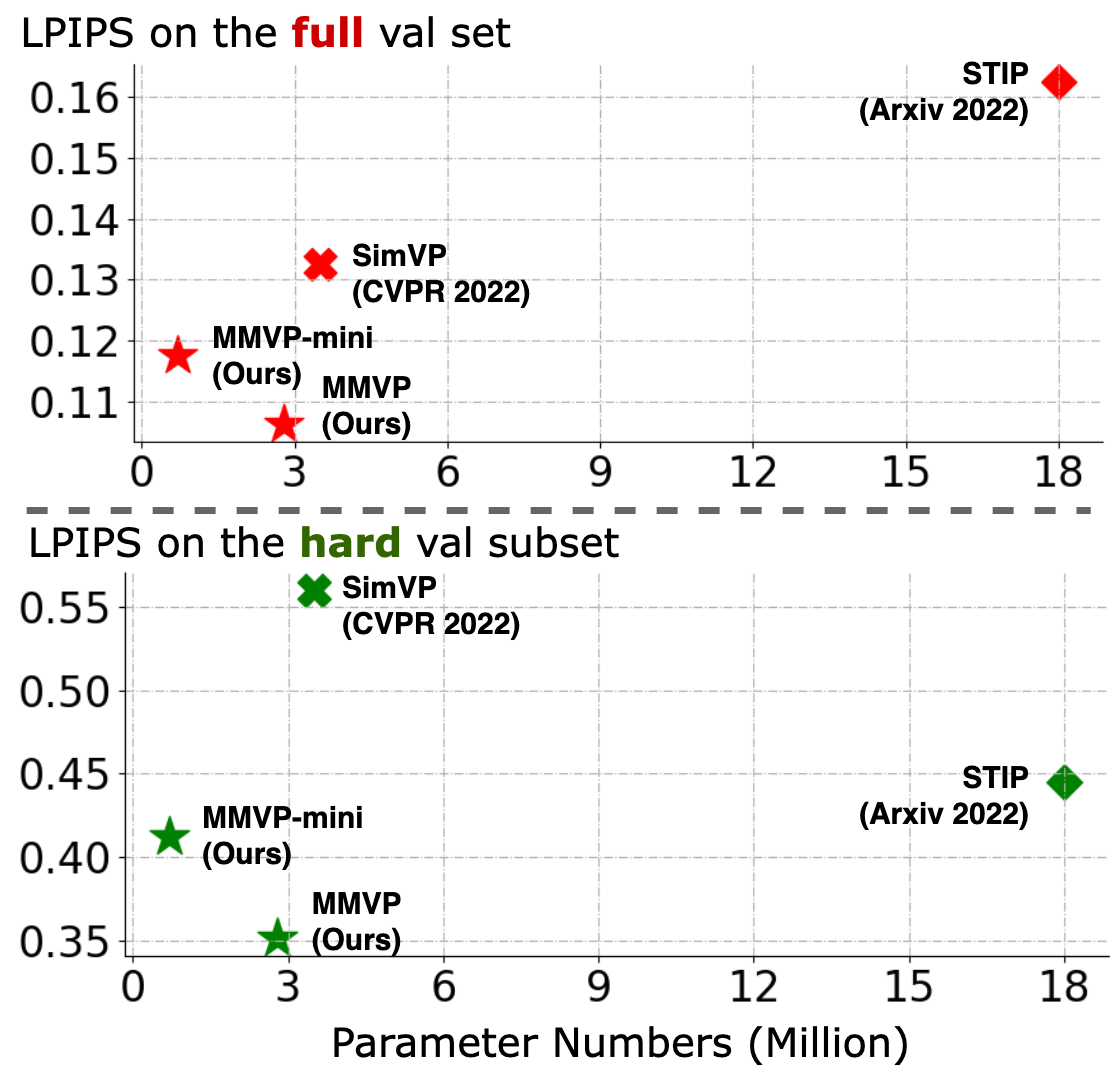}
    \vspace{-2mm}
    \caption{Performance comparison on UCF Sports with STIP~\cite{chang2022stip} and SimVP~\cite{gao2022simvp}. The \textit{hard} subset contains samples where SSIM between the last observed frame and the first future frame is smaller than 0.6, which indicates drastic motion patterns (data is from Tables~\ref{tab:ablation} and ~\ref{tab:ucf_result}). }
    \label{fig:table}
    \vspace{-2mm}
\end{figure}

\begin{figure*}[tbh]
    \centering
    \includegraphics[width=0.99\textwidth]{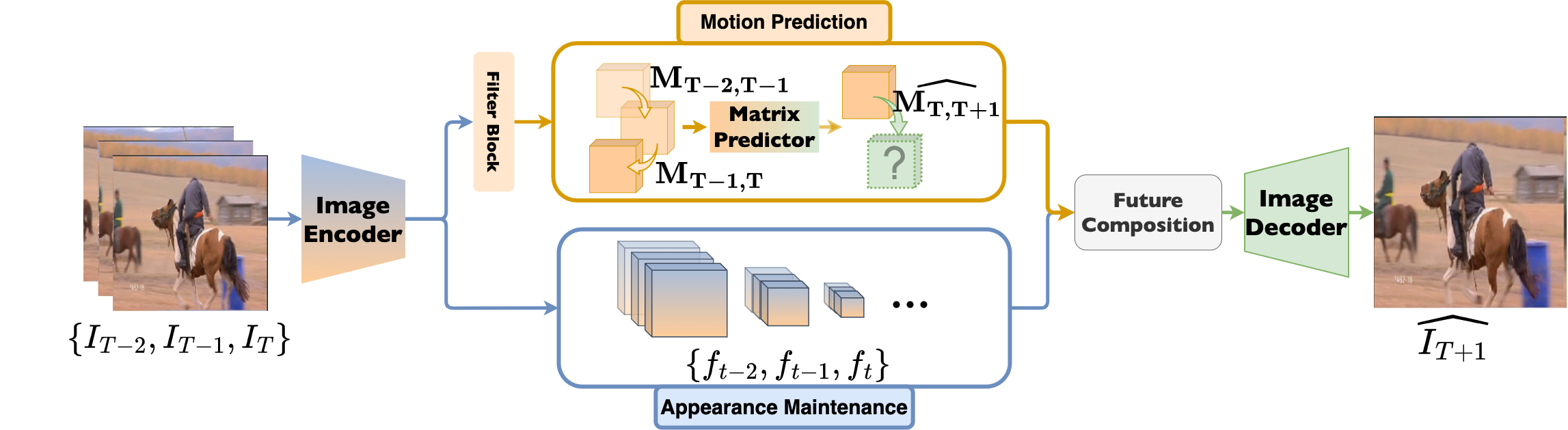}
     \vspace{-2mm}
    \captionof{figure}{MMVP is a two-stream video prediction framework. It decouples motion prediction and appearance maintenance, and it reunites motion and appearance features through feature composition operation.}
    \label{fig:pipeline}
    \vspace{-2mm}
\end{figure*}

Most existing works in video prediction are based on a single-stream pipeline that conducts motion prediction and appearance feature extraction for frame synthesis within the same set of modules. Their systems~\cite{wang2017predrnn,wang2018predrnn++,wang2018eidetic,yu2020efficient,chang2022strpm,chang2022stip} usually grow out of advanced network structures for sequential data analysis, such as recurrent neural networks (RNNs) \cite{mikolov2010recurrent} and transformers \cite{vaswani2017attention}. One shared characteristic of those methods is that they show excellent capabilities in capturing complex motion patterns but lower capabilities in appearance maintenance, yielding ``correct" but not ``good" synthesized frames. The reason behind this is that those methods usually contain complicated spatial-temporal feature extraction and state transition operations, which can cause unavoidable appearance information loss.

Researchers have proposed several solutions to mitigate appearance information loss; the most common approaches are introducing sophisticated appearance-aware state transition unit~\cite{chang2022strpm,chang2022stip,yu2019crevnet}, or adding frequent feature shortcuts from previous frames~\cite{gao2022simvp,wang2018eidetic,wang2017predrnn,wang2018predrnn++}. However, the former solution tends to build cumbersome models with a huge number of parameters; and for the latter solution, too much residual information from previous frames can cause a larger performance drop for hard cases such as videos with fast movements and/or moving cameras. Figure~\ref{fig:table} shows the comparison between STIP~\cite{chang2022stip} (an example of the former solution) and SimVP~\cite{gao2022simvp} (an example of the latter solution).

To avoid running into possible trade-offs between motion and appearance in single-stream pipelines, a few works have explored two-stream pipelines~\cite{liang2017dual,bei2021learning,gao2019disentangling,villegas2017decomposing,gao2021accurate}, decoupling motion prediction and appearance maintenance. However, they either require auxiliary sub-networks such as optical flow estimator~\cite{liang2017dual,wu2022optimizing} and key point detector~\cite{gao2021accurate} to generate motion representations, which complicates video prediction and reduces the generalizability of systems; or they do not provide an efficient solution to reunite the predicted motion and the appearance features~\cite{denton2017unsupervised,villegas2017decomposing}. 

\begin{figure*}[tbh!]
    \centering
    \includegraphics[width=0.99\textwidth]{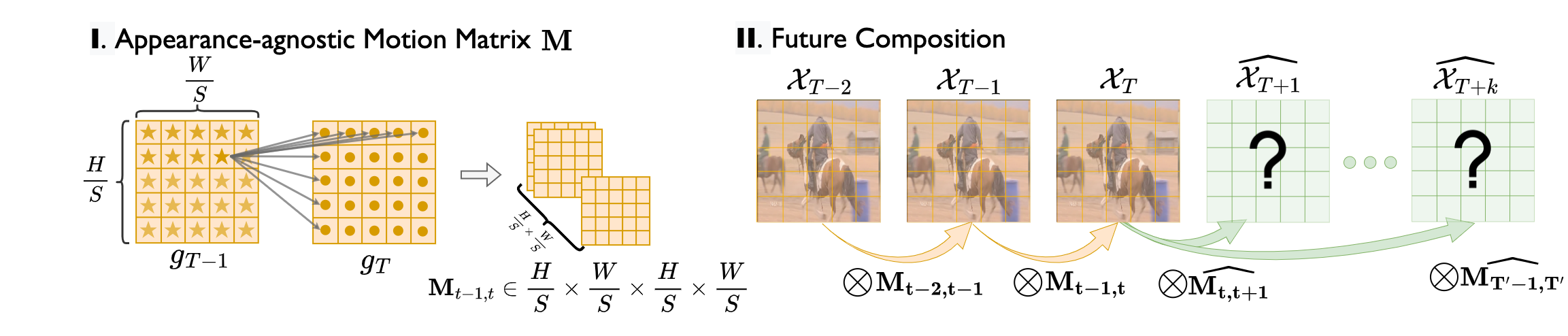}
     \vspace{-5mm}
    \captionof{figure}{MMVP relies on the motion matrix $\mathbf{M}$. $\mathbf{M}$ is an appearance-agnostic motion representation that measures the cosine similarity between the feature patches of two consecutive frames. Information of the past frames $\{\mathcal{X}_0,...,\mathcal{X}_T\}$ can compose the future information $\{\widehat{\mathcal{X}}_{T+1},...,\widehat{\mathcal{X}_{T'}}\}$ through matrix multiplication $\bigotimes$ with $\mathbf{M}$. }
    \label{fig:construction}
   \vspace{-5mm}
\end{figure*}

With these gaps in current research, we introduce a novel two-stream, end-to-end trainable framework for video prediction: Motion-Matrix-based Video Prediction (MMVP) (see Figure~\ref{fig:pipeline} for the framework overview). As the name indicates, MMVP uses motion matrices as the decoupled motion representation of video frames. The motion matrix is a 4D matrix representing the image feature patches of consecutive frames (see Figure~\ref{fig:construction} I). As motion matrices are the sole input of the matrix predictor (i.e., the motion prediction module in MMVP), MMVP specifies the hidden motion information and makes the matrix predictor only focus on motion-related information. For the reunion of motion and appearance features, MMVP gets inspiration from the image autoencoder. It first embeds the past frames individually through an image encoder. Then it composes the embedding of the future frames using the predicted motion matrices output by the matrix predictor and the past frames' embeddings through matrix multiplication~(see Figure~\ref{fig:construction} II). Then, an image decoder decodes the composed embeddings into the predicted future frames.



The advantage of MMVP is three-fold: (i) MMVP decouples motion and appearance by constructing motion matrices, requiring no extra construction modules; (ii) unlike optical flow that describes the one-to-one relationship between pixels, motion matrices describe the many-to-many relationship between feature patches, and are more flexible and applicable for real-world data; (iii) MMVP reunites the appearance and motion prediction results through matrix multiplication, which is interpretable and of little information loss. The advantages make MMVP a much more compact model with significantly fewer parameters yet still matching SOTA methods in performance. We validate MMVP on three datasets, UCF Sports~\cite{rodriguez2008action}, KTH~\cite{schuldt2004recognizing}, and MovingMNIST~\cite{srivastava2015unsupervised}. Experiments show that MMVP matches or surpasses SOTA methods on all three datasets across metrics. Specifically, compared to STIP~\cite{chang2022stip}, MMVP uses \textbf{84\%} fewer parameters (18M vs. 2.8M) but achieves \textbf{38\%} better performance in the LPIPS metric (12.73 vs. 7.88) on the UCF Sports dataset~(Table \ref{tab:ucf_strpm}).

\section{Related Works}
Good video prediction systems should not only accurately reason the future motions of the objects but also maintain their appearances and synthesize consistent future frames. As the video resolution has become increasingly higher today, researchers should additionally consider the scalability and efficiency of video prediction systems \cite{oprea2020review}.

Most video prediction works to date adopt a single-stream pipeline that grows out of advanced techniques in sequential data analysis. Those techniques usually contain sophisticated spatial-temporal feature extraction operations and state transitions, which result in appearance information loss. Thus, researchers tend to modify the techniques to let them attend more to appearance maintenance~\cite{wang2017predrnn,shi2015convolutional,wang2018predrnn++,chang2022strpm,chang2022stip,wang2018eidetic,hsieh2018learning}. Based on ConvLSTM~\cite{shi2015convolutional}, PredRNN~\cite{wang2017predrnn} propose a novel memory state transition method between the recurrent nodes to help the appearance features from observations be delivered to the final prediction. E3D-LSTM~\cite{wang2018eidetic} integrated 3D convolution operations to help emphasize the short-term appearance. CrevNet~\cite{kalchbrenner2017video} uses a conditionally reversible network to preserve the information from the past frames. More recently, the progress in temporal modeling made by transformer-based methods~\cite{vaswani2017attention} has drawn more attention. The field starts to see works that use transformers in video prediction tasks~\cite{gupta2022maskvit,weissenborn2019scaling,rakhimov2020latent}. These works have a better information aggregation capability and less appearance information loss, but they usually have a huge number of parameters and lack scalability to higher-resolution videos. 

Seeking an alternative solution to video prediction and overcoming the drawbacks of the aforementioned methods, a few works have also explored the potential of decoupling motion and appearance and building a two-stream pipeline. Using optical flow is an intuitive idea. Several previous works~\cite{liang2017dual,bei2021learning,wu2022optimizing,gao2019disentangling} use built-in or off-the-shell systems to predict the optical flow from the past to future frames; they use the predicted optical flow to warp past frames into final results. The drawbacks of this line of work are two-fold: i) time-consuming, especially for the warping procedure of high-resolution video sequences; ii) more importantly, the one-to-one relationship defined by the optical flow may not be applicable for some cases such as when one pixel or superpixel moves half of the pixel size, or when one pixel has impacts on several pixels in the next frames. Besides optical flow, pose~\cite{villegas2017decomposing} and keypoint~\cite{gao2021accurate} are also used to represent the motion of the objects in the scene for video prediction. However, this approach requires extra pose or keypoint detectors to process videos. For videos with multiple objects, more complex motions, and higher resolution, the efficiency of these extra modules cannot be guaranteed. Alternatively, MCNet~\cite{denton2017unsupervised} uses the difference between frames to represent motion information. It does not require extra modules, but using concatenation to reunite motion and appearance features is inefficient.

Compared to the small number of two-stream video prediction systems above, our MMVP has three advantages: i) its motion matrix is able to describe the many-to-many relationship of the pixels or super-pixels, which makes MMVP much more flexible and reasonable for real-world data; ii) MMVP does not require an extra module to produce motion representation; and iii) the reunion of motion and appearance features is more intuitive and interpretable than the approaches used in previous works. This paper will demonstrate how MMVP makes a more capable and compact video prediction framework.

\section{Motion-Matrix-based Video Prediction}
\subsection{Framework Overview}
Given a video sequence $\mathcal{I}=\{I_t\}_{t=1}^{T}$, where $I_t$ denotes the $t$th frame, usually in the RGB format, MMVP estimates the future $T'$ frames $\widehat{I_{T+1}}$ to $\widehat{I_{T+T'}}$ from $I_1$ to $I_T$. In comparison to the known frame set $\mathcal{I}$, we denote the estimated frame set as $\mathcal{I'}=\{I'_t\}_{t=T+1}^{T+T'}$. The training of the framework is solely supervised by mean-squared error (MSE) loss. MMVP consists of three steps as follows:


\textbf{Step 1: Spatial feature extraction.} We use an RRDBs~\cite{wang2018esrgan} based network architecture to model an image encoding function $\Omega$, defined in eq.~\ref{eq:encoder}. This image encoder embeds each frame $I_i, i\in\{1,2,...,T\}$ in a down-sample hidden space separately and outputs their corresponding hidden features $f_i$. In MMVP, besides the image encoder, we have an extra filter block that models a filtering function $\Theta$ defined in eq.~\ref{eq:trans}. It takes $f_i$ as the input and aims to generate $g_i$. As the module name indicates, the filter block aims to filter out motion-irrelevant features of $f_i$ for motion matrix construction. We will introduce the details in Sec.~\ref{sec:encoder}.

\textbf{Step 2: Motion matrix construction and prediction.} MMVP generates a set of motion matrices $\{\mathbf{M}_{i,i+1}\}, i\in\{1,2,..,T-1\}$ for every two consecutive frames based on their feature pairs $\{g_i,g_{i+1}\}$. Then, a matrix predicting function $\Phi$ defined in eq. \ref{eq:mat_pred} takes $\{\mathbf{M}_{i,i+1}\}$ as the input and predicts future matrices $\{\widehat{\mathbf{M}_{T,T+j}}\},j\in\{T,...,T+T'\}$. Sec.~\ref{sec:matrix} will elaborate on the definition and construction procedures of the motion matrices, and Sec.~\ref{sec:predictor} will demonstrate the inner structure of the matrix predictor.

\textbf{Step 3: Future composing and decoding.} Using the output of steps 1 and 2, MMVP composes the unknown information for future frames. Then, a future decoding module takes the composed features as the input and outputs the final prediction. We will introduce the feature composition procedure in Sec. \ref{sec:feat_compose} and the future decoding module in Sec. \ref{sec:feat_decode}.

\subsection{Spatial Feature Extraction}
\label{sec:encoder}
Spatial feature extraction involves two components of the MMVP framework: image encoder, and filter block. 

The image encoder $\Omega$ in MMVP encodes every $I_i$ from the input data sequence to their corresponding features $f_i$ individually. The filter block $\Theta$ subsequently processes $f_i$ and makes it ready for the construction of a motion matrix. Formally,
\begin{equation}
 \vspace{-3mm}  
    f_i = \Omega(I_i),
    \label{eq:encoder}
\end{equation}
 \vspace{-2mm}  
\begin{equation}
    g_i = \Theta(f_i),
    \label{eq:trans}
\end{equation}
where $i\in \{1,2,...,T\}$. We use a convolutional network with residual in residual dense blocks (RRDBs)~\cite{wang2018esrgan} to implement the image encoder, and we use a two-layer convolutional network to implement the filter block. See the detailed network architecture in Appendix. 

Next, the output features of the image encoder will take participate in the future feature composition, while the output of the filter block is only used for motion matrices construction. The existence of the filter block helps the model to filter out irrelevant features from the image encoder output and allows the construction of motion matrices to focus more on motion-related features. See Table \ref{tab:ablation} for the ablation study about the filter block.

\subsection{Motion Matrix Construction}
\label{sec:matrix}
Given the output of the filter block $g_{i}\in \mathbb{R}^{\frac{H}{S}\times \frac{W}{S}},i
\in\{1,2,...,T\}$, we denote the feature patch at $(h,w),w\in \{0,1,2,\frac{W}{S}-1\},h\in \{0,1,2,\frac{H}{S}-1\}$ as $g_i^{h,w}$, where $H$ and $W$ are the input images' height and width, and $S$ is the downsampling ratio to the original image; thus $\frac{H}{S}$ and $\frac{W}{S}$ are respectively the feature map's height and width.

For two consecutive frames' feature $\{g_{i},g_{i+1}\}$, we calculate the cosine similarity for each and every pair of feature patches to construct a 4D motion matrix $\mathbf{M}_{i,i+1}\in \mathbb{R}^{\frac{H}{S}\times \frac{W}{S}\times \frac{H}{S}\times \frac{W}{S}}$. We denote the element of the matrix $\mathbf{M}_{i,i+1}$ at ${(h_i,w_i,h_{i+1},w_{i+1})}$ as $\mathbf{M}_{i,i+1}^{h_i,w_i,h_{i+1},w_{i+1}}$, and let

\begin{equation}
    \mathbf{M}_{i,i+1}^{h_i,w_i,h_{i+1},w_{i+1}} = D_c(g_i^{h_i,w_i},g_{i+1}^{h_{i+1},w_{i+1}}).
\end{equation}In the equation, $D_c$ is the cosine similarity, $g_i^{h_i,w_i}$ is the feature patch of frame $i$ with the index of ${(h_i,w_i)}$, and $g_{i+1}^{h_{i+1},w_{i+1}})$ is the feature patch of frame $i+1$ with the index of ${(h_{i+1},w_{i+1})}$.
With a feature patch at $(h_i,w_i)$ of $g_i$ as an example, the matrix $\mathbf{M}_{i,i+1}^{h_i,w_i}\in \mathbb{R}^{H\times W}$ can be regarded as a heatmap that reflects how much impact $g_i^{h_i,w_i}$ has on $g_{i+1}$, or more intuitively, the motion tendency of $g_i^{h_i,w_i}$ (see Figure~\ref{fig:matrix} for illustration). It is similar to the definition of optical flow~\cite{horn1981determining}, which determines the movement of a pixel or superpixel. The difference between optical flow and motion matrix is that motion matrix does not strictly define a one-to-one relationship between the pixels or superpixels of the current frame and those of the next frame. It is a more practical assumption in the video prediction task that one current feature patch may influence several future feature patches. 

\subsection{Matrix Prediction}
\label{sec:predictor} 
Given the motion matrices of the past $T$ frames $\{\mathbf{M}_{1,2},\mathbf{M}_{2,3},...,\mathbf{M}_{T-1,T}\}$, the matrix predictor $\Psi$ predicts the future matrices $\{\mathbf{M}_{T,T+1},\mathbf{M}_{T,T+2},...,\mathbf{M}_{T,T'}\}$. Instead of predicting the motion matrices between the consecutive frames, we predict the ones between the last observed frame $I_T$ and every future frame $I_{T+j},j\in\{1,2,...,T'\}$, as 
\begin{equation}
    \{\widehat{\mathbf{M}_{T,{T+j}}^{w,h}}\} = \Psi(\{\mathbf{M}_{i,{i+1}}^{w,h}\}),\forall i\in\{1,2,...,T-1\},
    \label{eq:mat_pred}
\end{equation}
where $\mathbf{M}_{i,{i+1}}^{w,h} \in \mathbb{R}^{H\times W}$, $w\in \{1,2,...,\frac{W}{S}\}$ and $h\in \{1,2,...,\frac{H}{S}\}$.
This design aims to reduce the accumulative error during feature composition and is validated by the long-term prediction setting shown in Table~\ref{tab:kth_result}. Since for this work, we focus on testing the function and performance of the MMVP framework, we report the use of a simple \emph{3D fully convolutional} architecture to implement $\Psi$. Fellow researchers can choose to easily replace the implementation of $\Psi$ with more advanced temporal modules if they wish to pursue better performances.

\begin{figure}[tbh!]
    \centering
    \includegraphics[width=0.48\textwidth]{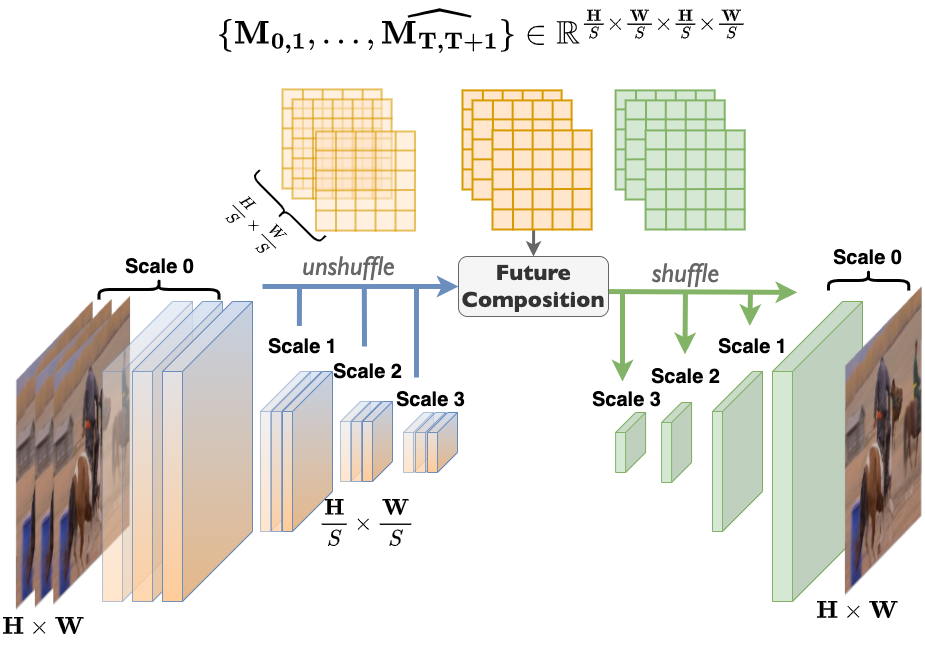}
     \vspace{-3mm}  
    \caption{All scales of image features and the original frames can join the future composition with predicted motion matrices through pixel unshuffle and shuffle operations.} 
    \label{fig:multiscale}
     \vspace{-3mm}  
\end{figure}
\subsection{Multi-scale Future Composition}
\label{sec:feat_compose} 
The future composition step generates information for the future frames using the observed information and the motion matrices. It is formulated as  
\begin{equation}
    \widehat{\mathcal{X}_{T+j}} = \sum_{i=1}^{T}(\mathcal{X}_i \times \prod_{n=i}^{T-1}\mathbf{M}_{n,n+1} \times \widehat{\mathbf{M}_{T,T+j}}).
    \vspace{-1mm}
\end{equation}
As the equation indicates, instead of only using the information of the last observed frame, we use all observed information for future composition and lower the weight of earlier frames through repeated multiplication of the motion matrices. The $\mathcal{X}$ in the equation represents the observed information of the past frames. The information can be the output features of the image encoder with different scales $f_i\in \mathbb{R}^{H^s\times W^s\times C}$, where $C$ is the feature length; the information can also be the observed frames $I_i\in \mathbb{R}^{H\times W\times 3}$. Since the motion matrices are constructed from a certain scale of image features, there will be incompatibilities between the matrices and some features. To enable matrix multiplication between the motion matrices and the observed features (at any scales) or images, we borrow the pixel unshuffle and shuffle operations from \cite{shi2016real}. The pixel unshuffle operation reshapes features or images into the identical scale as the motion matrices for matrix multiplication. Afterward, the pixel shuffle operation reshapes the results of the matrix multiplication back to the original scales of the features or images. See Figure \ref{fig:multiscale} for demonstration. This whole process involves little information loss. In Table~\ref{tab:ucf_result}, we examine the multi-scale feature composition design. We find out that in general, the system achieves better performance when more scales of features take part in the feature composition.

\subsection{Future Decoding}
\label{sec:feat_decode}
The future decoding procedure is used for aggregating and processing the composed features to formulate the final output. Since in the feature composition procedure, all scales of features are able to compose their corresponding scales of features for future frames, we adopt the decoder structure of UNet~\cite{ronneberger2015u} with RRDB blocks~\cite{wang2018esrgan} in the end to implement MMVP's image decoder. This design allows the composed features from all scales of image features, as well as the original images, to contribute to the final output. We use MSE loss to supervise the framework training.

\section{Experiment}
\label{sec:exp}
In this section, we evaluate MMVP quantitatively and qualitatively against existing state-of-the-art methods on various publicly available datasets. In addition, we analyze MMVP through a set of ablation studies on the UCF Sports dataset to better illustrate our design logic.

\subsection{Datasets}
We briefly review three widely used datasets and their configurations for MMVP evaluation.
\begin{table}[tbh]
 \vspace{-2mm}  
\caption{Experiment settings for each dataset.}
     \vspace{-2mm}  
    \centering
    \scalebox{0.85}{
    \begin{tabular}{c|ccc}
    \toprule
    Dataset	&	 Resolution	& Train	& Test \\
    \hline
    \hline
    UCF Sports     & 	$512\times512$	&	$ 4 \rightarrow 1$	&	$ 4 \rightarrow 6$	\\
    KTH     & 	$128\times128$	&    $10 \rightarrow 20$ or $40$  &	$ 10 \rightarrow 20$ or $40$	\\
    Moving MNIST     & 	$64\times64$	&	$ 10 \rightarrow 10$	&	$ 10 \rightarrow 10$	\\
    \toprule

    \end{tabular}}
    \label{tab:exp_set}
     \vspace{-2mm} 
\end{table}

\textbf{UCF Sports~\cite{rodriguez2008action}} contains 150 video sequences collected from various sports scenes with 10 different action types including running, kicking, and diving. It is regarded as a challenging dataset given its low frame rate (10 fps), high resolution ($480 \times 720$), and complex motion patterns. The video sequences in this dataset contain both rapid foreground (e.g. athletes) and background movements (e.g. the camera motion). We perform two different training/validation splits of this dataset, splits from STRPM \cite{chang2022strpm} and our own splits, to facilitate different evaluation purposes. We use the STRPM splits for performance comparison with other methods (Table~\ref{tab:ucf_strpm}). In STRPM, long video sequences are cut into several short clips; for a particular sequence, some of its clips may be put into the training split, and others into the validation split.
To avoid the appearance features of certain sequences in the validation set to be exposed in the training set, and to thoroughly evaluate MMVP's capability of video prediction, we generate our own splits for UCF Sports, which choose 90\% of the video sequences from each action type to form the training samples, and the rest 10\% to form the validation samples. 

Notably, within the same validation set, the difficulty level of different samples varies a lot. Some easier video clips contain static backgrounds and slow-moving objects, while others are more difficult for involving drastic camera movement and/or fast-moving objects. To better understand the model's prediction ability in different scenarios, we apply certain thresholds of the structural similarity index measure (SSIM) between the last observed frame and the first future frame to divide the UCF Sports validation set into three subsets: the easy (SSIM $\leq$ 0.9), intermediate (0.6 $\leq$ SSIM $<$ 0.9), and hard subsets (SSIM $<$ 0.6),
which respectively take up 66\%, 26\%, and 8\% of the full set.

We use our own UCF Sports splits in the ablation studies (Sec.~\ref{sec:ablation_study}). For comparisons, we also train and test STIP~\cite{chang2022stip} (extension of STRPM~\cite{chang2022strpm}) and SimVP~\cite{gao2022simvp} on our own UCF Sports splits; see Table~\ref{tab:ucf_result}.

\textbf{KTH~\cite{schuldt2004recognizing}} contains videos of 25 individuals performing six types of actions, i.e., walking, jogging, running, boxing, hand-waving, and hand-clapping. Following previous works~\cite{villegas2017decomposing,wang2018eidetic,gao2022simvp}, we use persons 1-16 for training and persons 17-25 for validation. The videos in KTH are in grayscale, but the challenging part of this dataset is its experiment setting, which requires the system to output 20 or 40 frames given only 10 past frames.

\textbf{Moving MNIST~\cite{srivastava2015unsupervised}} is a synthetic dataset. Each sequence in the dataset consists of two digits moving independently within the $64 \times 64$ grid and bouncing off the boundary. During the training time, by assigning different initial locations and velocities to each digit, one can generate an infinite number of sequences, and train the system to predict the future 10 frames from the previous 10 frames. For fair comparisons, we use the pre-generated 10000 sequences\cite{gao2022simvp} for validation.

\subsection{Metrics}
We use the peak signal-to-noise ratio~(\textbf{PSNR}) and structural similarity index measure~(\textbf{SSIM}) to evaluate the image quality of the predicted frames. We use the implementation of PSNR and SSIM from the package \textit{scikit-learn}~\cite{scikit-learn}. For the UCF Sports dataset, we use Learned Perceptual Image Patch Similarity~(\textbf{LPIPS})~\cite{zhang2018unreasonable} for better comparison with existing methods. LPIPS represents the perceptual quality of the predicted frames. It measures the feature distance of corresponding image patches between the predicted frames and the ground truth. Following the setting of these previous works, in the experiments on Moving MNIST, we use the sum of the mean squared error~(\textbf{MSE}) from the entire frame to evaluate the image quality.

\begin{figure*}[tbh!]
    \centering
    \includegraphics[width=0.85\textwidth]{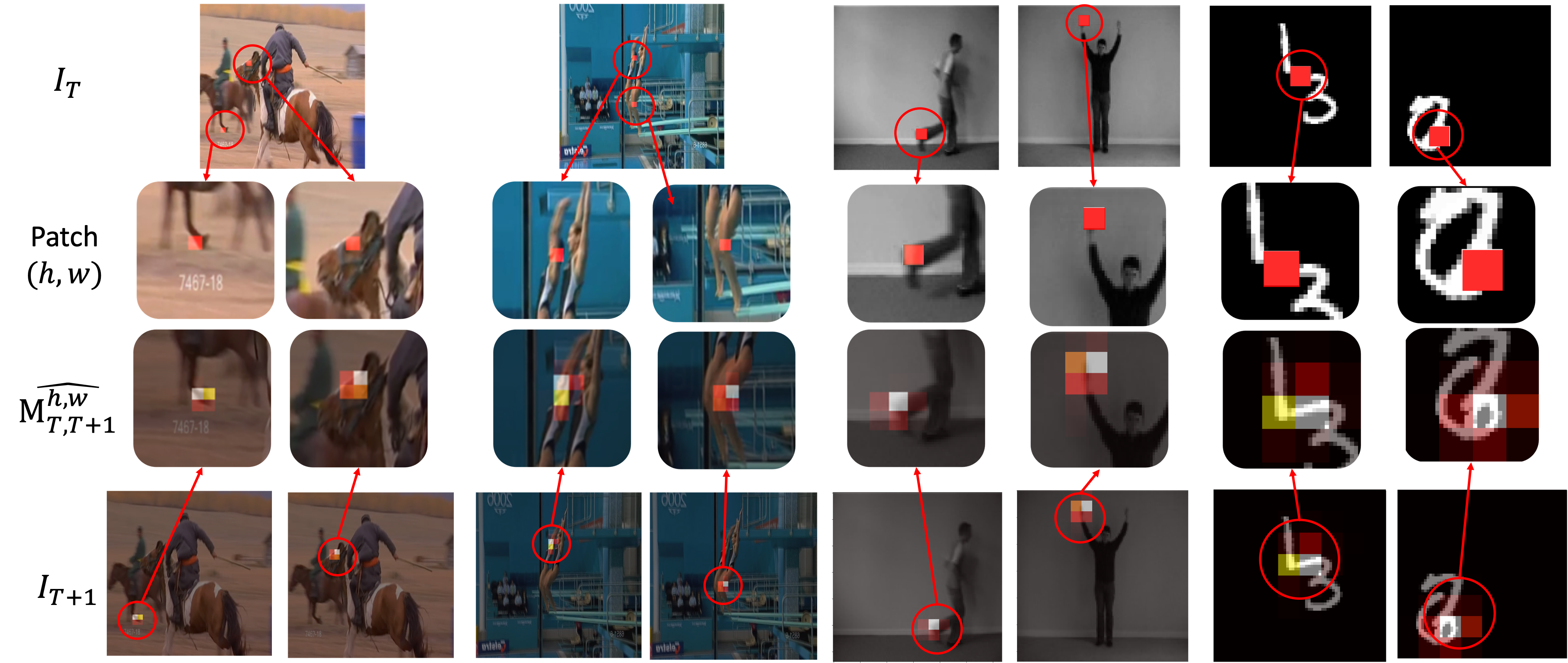}
    \caption{Predicted motion matrix visualization. We highlight the selected feature patch(es) at $(h,w)$ in the last observed frame $I_{T}$ in red and visualize their corresponding predicted motion matrices ${\mathbf{M}_{T,T+1}^{h,w}}\in \mathbb{R}^{H\times W} $ overlaying with the first future frame $I_{T+1}$. A brighter color indicates a higher predicted value. We select two samples for each dataset. From left to right, samples originate in the validation set of UCF Sports, KTH, and Moving MNIST. }
    \label{fig:matrix}
\vspace{-2mm}
\end{figure*}

\subsection{Ablation Study}
\label{sec:ablation_study}
We conduct the ablation study of future composition~(Table~\ref{tab:ucf_result}), filter block~(3rd and 4th rows of Table~\ref{tab:ablation}), and hyperparameters for both the image encoder/decoder and the matrix predictor (1st, 2nd, and 4th rows of Table~\ref{tab:ablation}) on our splits of the UCF Sports dataset. 

We first examine how the number of feature scales for feature composition impacts the results. Since we use the UNet structure with three times downsampling/upsampling operations to encode, we have four different scales of features for composition. For UCF Sports, the scales are $1, \frac{1}{2}, \frac{1}{8}, \frac{1}{16}$. The motion matrices are constructed using features of $\frac{1}{8}$ scale. From Table~\ref{tab:ucf_result} 3rd to 5th rows, we observe consistent performance boosts on all metrics and subsets when we involve more scales of features for composition. We then test if using images for feature composition also boosts performance. By comparing the 5th and 6th rows of Table~\ref{tab:ucf_result}, we can still see an increasing pattern. However, since the scale of the image is identical to the largest scale of the features, to avoid redundancy and possible conflicts, we remove the features of the largest scale and produce the results of the last row in Table~\ref{tab:ucf_result}. We then see a comparable result to the one that uses all scales of features as well as the image. We adopt the model setting of the last row for the rest of the experiments.  

As mentioned in Sec.~\ref{sec:encoder}, we do not directly use the output of the image encoder to construct the motion matrices. Instead, we add a filter block to filter out irrelevant features and help the motion matrices focus more on describing the temporal similarity regarding the motion-related features. Table~\ref{tab:ablation} shows that the model with the filter block generally achieves better results on the UCF Sports validation set.

  

\begin{table}[tbh!]
    \centering
  \caption{Ablation study on UCF Sports with LPIPS metrics (the lower the better), including the feature length of image and motion, and the usage of the filter block (F-Block).}  
    \label{tab:ablation}
    \scalebox{0.72}{\begin{tabular}{ccc|cccc|c}
      \toprule
Img  &   Motion     &   F-Block    &  Full &   Easy    &   Intermediate    &   Hard    &   Param \#\\
\hline
\hline
16  &   4     & $\checkmark$&           0.1184    &  0.0592   &   0.1811   &  0.4168  &   0.70M\\
16  &   8     & $\checkmark$&       0.1175    &   0.0600   &   0.1768   &  0.4122   &   0.71M\\

32  &   8     &  $\times$   &       {0.1124}   &  \textbf{0.0574}  &  {0.1729}  &   {0.3819} &  2.57M\\
32  &   8     &$\checkmark$ &   {\textbf{0.1062}}    &    {0.0580}   &    {\textbf{0.1569}}  &    {\textbf{0.3510}} &  2.79M\\
\hline
\hline
\multicolumn{3}{c|}{STIPHR~\cite{chang2022stip}}  &   0.1626  &   0.1066  &   0.2271  &   0.4450  &   18.05M\\
\multicolumn{3}{c|}{SimVP~\cite{gao2022simvp}}  &   0.1326  &   0.0584  &   0.1951  &   0.5600  &   3.47M  \\

    \toprule
    \end{tabular}
    }  
     \vspace{-3mm} 
\end{table}
Furthermore, Table~\ref{tab:ablation} also shows an ablation study on the feature length of the appearance-related modules (image encoder and decoder) and the motion-related module (matrix predictor). We define the configuration of the second row as the MMVP-mini for Figure~\ref{fig:table}.

\subsection{Motion Matrix Visualization}
In Figure~\ref{fig:matrix}, we visualize the predicted motion matrices for certain selected feature patches. In all the demonstrated samples, the many-to-many relationships described by the predicted motion matrices are able to accurately capture the feature patches moving tendency. They show that the motion matrix can describe a wide range of motion patterns and scenarios, including multiple objects (1st sample of UCF Sports), the single person moving~(KTH), multiple persons moving and camera moving (2nd sample of UCF Sports), and synthesized motions (Moving MNIST).

\subsection{Comparison with SOTA Methods}

We compare MMVP with existing SOTA methods on three popular datasets: UCF Sports (STRPM~\cite{chang2022strpm} splits, Table~\ref{tab:ucf_strpm}), KTH (Table~\ref{tab:kth_result}),   and Moving MNIST (Table~\ref{tab:mnist_result}). 

On the UCF Sports dataset, we observe large performance gains by MMVP compared to other methods. As mentioned above, the motion patterns in UCF Sports are the most complex of the three datasets due to many difficult cases, e.g., fast movement, camera moving, and motion blur. But the two-stream design of MMVP has shown its ability in such complex scenarios. Meanwhile, the video resolution of UCF Sports is also the highest among the three datasets. Most existing methods in Table~\ref{tab:ucf_strpm} are inherently not designed for high-resolution videos, which requires larger network capacities to maintain appearance information. Although STRPM~\cite{chang2022strpm} and STIP~\cite{chang2022stip} proposed residual temporal modules designed for high-resolution videos and they could largely surpass previous methods on both metrics, the appearance information loss in their methods is still unavoidable. MMVP, nevertheless, allows the image and multi-scale features to reach the decoder through a feature composing procedure, successfully minimizes the information loss, and achieves the best performance among all methods across all metrics. MMVP's success on the UCF Sports dataset validates its readiness for real-world applications, and its scalability for high-resolution videos. 

Another notable result is MMVP's ability in long-term information preservation; see Table~\ref{tab:kth_result}. Despite having lower video resolutions and less complex motion patterns than UCF Sports, KTH as a real-world dataset still presents a difficulty as it requires systems to do long-term prediction given short-term observations. On the KTH dataset, MMVP achieves the \#1 performance on SSIM metrics for both experiment settings with a small performance gap (27.54 to 26.35) in PSNR between the two settings. This is contributed by the design that the matrix predictor of MMVP predicts the temporal similarity matrices between the future frames and the last observed frame instead of their predecessors. Then, every predicted frame is composed of valid observed information. It reduces accumulated errors and ensures good performance for long-term prediction.

Compared to the performance gains on UCF Sports, MMVP does not show much advantage on the Moving MNIST dataset. One speculation is that with video prediction research advancing, the problem of two-digits Moving MNIST with low resolution ($64\times 64$) is nearly solved. With sufficient training time~\cite{gao2022simvp}, current video prediction systems may all be able to achieve promising results. To further promote the growth of this field, researchers can consider more challenging datasets and experiment settings. 

\begin{table}[tbh!]
\caption{Performance comparison on the KTH dataset}
\vspace{-2mm}    
    \centering \small
    \resizebox{\linewidth}{!}{%
    \begin{tabular}{ccccc}
    \toprule
    \multirow{2}{*}{Method}   &   \multicolumn{2}{c}{KTH$ 10 \rightarrow 20$}   & \multicolumn{2}{c}{KTH$ 10 \rightarrow 40$} \\

       &  SSIM $\uparrow$ & PSNR $\uparrow$ & SSIM $\uparrow$ & PSNR $\uparrow$\\
       
       \hline
       \hline
MCnet~(ICLR2017)~\cite{villegas2017decomposing}	&	0.804	&	25.95	&	0.73	&	23.89 \\
ConvLSTM~(NeurIPS2015)~\cite{shi2015convolutional}	&	0.712	&	23.58	&	0.639	&	22.85 \\
SAVP~(arXiv2018)~\cite{lee2018stochastic}	&	0.746	&	25.38	&	0.701	&	23.97 \\
VPN~(PMLR2017)~\cite{kalchbrenner2017video}	&	0.746	&	23.76	&	–	&	– \\
DFN~(NeurIPS2016)~\cite{jia2016dynamic}	&	0.794	&	27.26	&	0.652	&	23.01 \\
fRNN~(ECCV2018)~\cite{oliu2018folded}	&	0.771	&	26.12	&	0.678	&	23.77 \\
Znet~(ICME2019)~\cite{zhang2019z}	&	0.817	&	27.58	&	–	&	– \\
SV2Pi~(ICLR2018)~\cite{babaeizadeh2017stochastic}	&	0.826	&	27.56	&	0.778	&	25.92 \\
SV2Pv~(ICLR2018)~\cite{babaeizadeh2017stochastic}	&	0.838	&	27.79	&	0.789	&	26.12 \\
PredRNN~(NeurIPS2017)~\cite{wang2017predrnn}	&	0.839	&	27.55	&	0.703	&	24.16 \\
VarNet~(IROS2018)~\cite{jin2018varnet}	&	0.843	&	28.48	&	0.739	&	25.37 \\
SVAP-VAE~(arXiv2018)~\cite{lee2018stochastic}	&	0.852	&	27.77	&	0.811	&	26.18 \\
PredRNN++~(ICML2018)~\cite{wang2018predrnn++}	&	0.865	&	28.47	&	0.741	&	25.21 \\
MSNET~(BMVC2019)~\cite{lee2018mutual}	&	0.876	&	27.08	&	–	&	– \\
E3d-LSTM~(ICLR2019)~\cite{wang2018eidetic}	&	0.879	&	29.31	&	0.810	&	27.24 \\
STMFANet~(CVPR2020)~\cite{jin2020exploring}	&	0.893	&	\textbf{29.85}	&	0.851	&	\textbf{27.56} \\
\hline
MMVP (ours)	&	\textbf{0.906}		&	27.54		& \textbf{0.888}			&	26.35		\\
    \toprule
    \end{tabular}
\label{tab:kth_result}}
 \vspace{-3mm}   
\end{table}

\begin{figure*}[tbh!]
    \centering
    \captionsetup{type=figure}
    \includegraphics[width=0.9\textwidth]{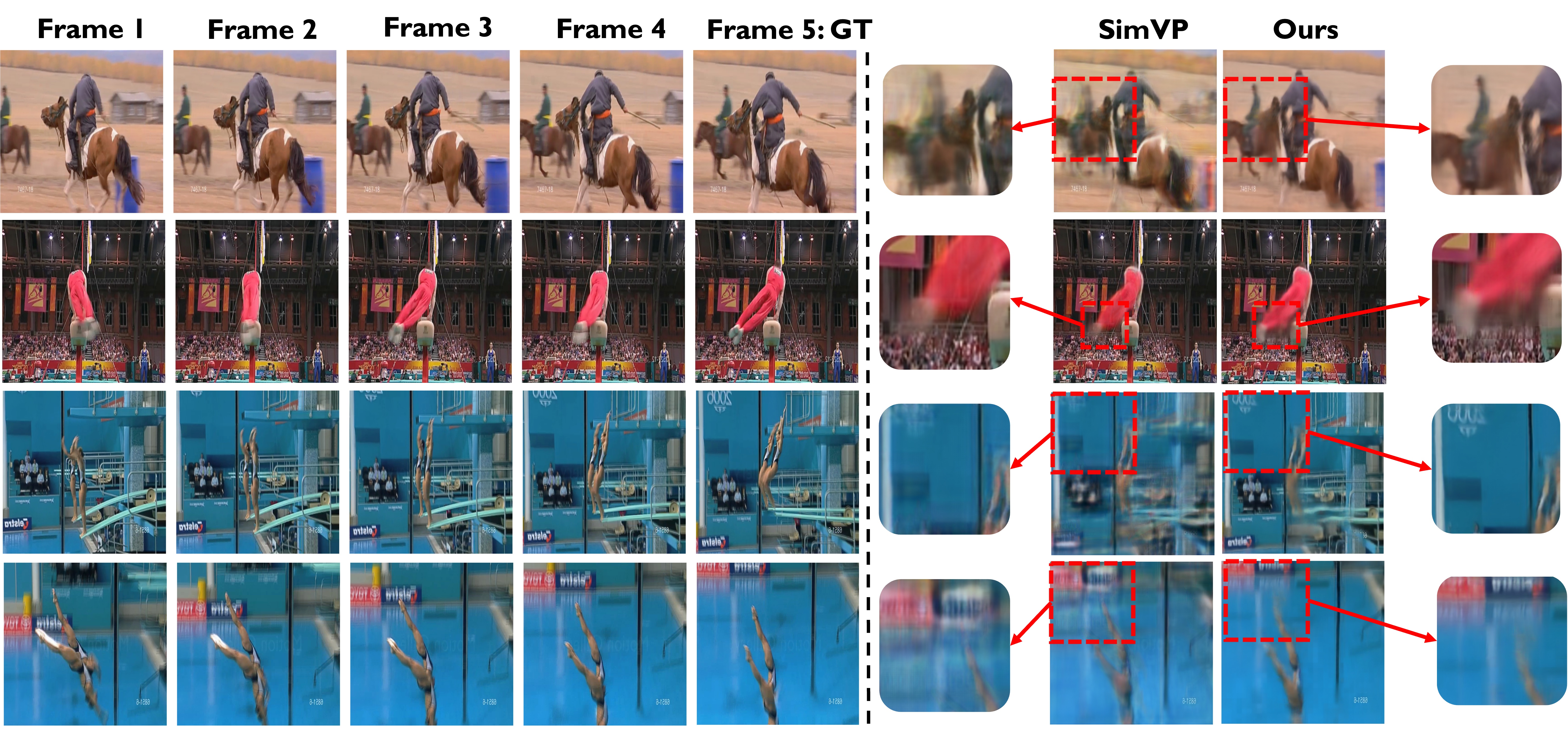}
\vspace{-2mm}
    \captionof{figure}{Qualitative results on our own splits of the UCF Sports dataset. }
    \label{fig:qual}
\vspace{-2mm}
\end{figure*}
\begin{table*}[tbh!]
\begin{minipage}{.55\linewidth}
\caption{Performance comparison on UCF Sports STRPM split}
\vspace{-2mm}    
    \centering \small
    \resizebox{\linewidth}{!}{%
    \begin{tabular}{ccccc}
    \toprule
    \multirow{2}{*}{Method}   &   \multicolumn{2}{c}{$t = 5$}   & \multicolumn{2}{c}{$t = 10$} \\

       &  PSNR $\uparrow$ & LPIPS$_{\times 100} \downarrow$ & PSNR $\uparrow$ & LPIPS$_{\times 100}$ $\downarrow$ \\
       
       \hline
       \hline
ConvLSTM (NeurIPS2015)~\cite{shi2015convolutional}	&	26.43	&	32.20	&	17.80	&	58.78 \\
BeyondMSE (ICLR2016)~\cite{mathieu2015deep}	&	26.42	&	29.01	&	18.46	&	55.28 \\
PredRNN (NeurIPS2017)~\cite{wang2017predrnn}	&	27.17	&	28.15	&	19.65	&	55.34 \\
PredRNN$++$ (ICML2018)~\cite{wang2018predrnn++}	&	27.26	&	26.80	&	19.67	&	56.79 \\
SAVP (arXiv2018)~\cite{lee2018stochastic}	&	27.35	&	25.45	&	19.90	&	49.91 \\
SV2P (ICLR2018)~\cite{babaeizadeh2017stochastic}	&	27.44	&	25.89	&	19.97	&	51.33 \\
E3D-LSTM (ICLR2019)~\cite{wang2018eidetic}	&	27.98	&	25.13	&	20.33	&	47.76 \\
CycleGAN (CVPR2019)~\cite{kwon2019predicting}	&	27.99	&	22.95	&	19.99	&	44.93 \\
CrevNet (ICLR2020)~\cite{yu2020efficient}	&	28.23	&	23.87	&	20.33	&	48.15 \\
MotionRNN (CVPR2021)~\cite{wu2021motionrnn}	&	27.67	&	24.23	&	20.01	&	49.20 \\
STRPM (CVPR2022)~\cite{chang2022strpm}	&	28.54	&	20.69	&	20.59	&	41.11 \\
STIP (arXiv2022)~\cite{chang2022stip}	&	30.75	&	12.73	&	21.83	&	39.67 \\
DMVFN (CVPR2023)~\cite{hu2023dynamic} &  30.05 & 10.24 & 22.67 &  22.50  \\
\hline
MMVP (Ours)	&	\textbf{31.68}	&	\textbf{7.88}	&	\textbf{23.25}	&   \textbf{22.24}	 \\
    \toprule
    \end{tabular}
}
    
    \label{tab:ucf_strpm}
\end{minipage}
\hspace{0.5cm}
\begin{minipage}{.4\linewidth}

    \caption{Comparisons on Moving MNIST}
    \vspace{-2mm}    
    \centering \small
    \scalebox{0.85}{
    \begin{tabular}{ccc}
    \toprule
Method	&	MSE	&	SSIM	\\
\hline
\hline
ConvLSTM~(NIPS 2015)~\cite{shi2015convolutional}	&	103.3	&	0.707	\\
PredRNN~(NIPS 2017)~\cite{wang2017predrnn}	&	56.8	&	0.867	\\
PredRNN-V2~(Arxiv 2021)~\cite{wang2022predrnn}	&	48.4	&	0.891	\\
CausalLSTM~(ICML 2018)~\cite{wang2018predrnn++}	&	46.5	&	0.898	\\
MIM~(CVPR 2019)~\cite{wang2019memory}	&	44.2	&	0.910	\\
E3D-LSTM~(ICLR 2018)~\cite{wang2018eidetic}	&	41.3	&	0.920	\\
PhyDNet~(CVPR 2020)~\cite{guen2020disentangling}	&	24.4	&	0.947	\\
CrevNet~(ICLR 2020)~\cite{yu2020efficient}	&	22.3	&	0.949	\\
SimVP~(CVPR 2022)~\cite{gao2022simvp}	&	23.8	&	0.948	\\
\hline
MMVP(ours)& \textbf{22.2}  &  \textbf{0.952}\\
    \toprule
    \end{tabular}
    \label{tab:mnist_result}
}
    \end{minipage}

\end{table*}

Moreover, we analyze MMVP's performance on data of different difficulty levels using our own split of the UCF Sports dataset. We define the difficulty of a video using the SSIM between the last observed frame and the first frame to be predicted. This is because a lower SSIM indicates a larger difference between the two frames, and less possibility for a system to take the shortcut by using the residual information from past frames and not actually predicting the motion. Table~\ref{tab:ucf_result} shows the quantitative evaluation compared with SimVP~\cite{gao2022simvp} and STIP~\cite{chang2022stip}: as the difficulty level of the validation subset increases, the performance gap between MMVP and other methods also increases. 

\begin{table*}[tbh!]
\caption{Ablation study on sources for future composition and the comparison with other SOTA methods on UCF Sports. }
\vspace{-2mm}
    \centering \small
    \scalebox{0.74}{
    
    \begin{tabular}{ccccccccccccccccccc}
    \toprule
    \multirow{2}{*}{Method}   &   \multicolumn{5}{c}{Composition source}  &   \multicolumn{3}{c}{Full set}   & \multicolumn{3}{c}{Easy (SSIM $\geq 0.9$)}   &   \multicolumn{3}{c}{Intermediate~(0.6 $\leq$ SSIM $<$ 0.9) }    &   \multicolumn{3}{c}{Hard (SSIM $<0.6$)}& \multirow{2}{*}{Param\#}  \\
    \cline{7-18}
       &Img&1&$1/2$&$1/8$&$1/16$& SSIM $\uparrow$ & PSNR $\uparrow$ & LPIPS $\downarrow$ & SSIM $\uparrow$ & PSNR $\uparrow$ & LPIPS $\downarrow$ & SSIM $\uparrow$ & PSNR $\uparrow$ & LPIPS$ \downarrow$ & SSIM $\uparrow$ & PSNR $\uparrow$ & LPIPS $\downarrow$& \\
       \hline
       \hline
    STIPHR~\cite{chang2022stip}  &    \multicolumn{5}{c}{-}	&	0.8817	&	28.17	&	0.1626	&	0.9491	&	30.65	&	0.1066	&	0.8351	&	23.97	&	0.2271	&	0.4673	&	15.97	&	0.4450 &   	18.05M\\	
    SimVP~\cite{gao2022simvp}  &    \multicolumn{5}{c}{-}	&	0.9189	&	29.97	&	0.1326	&	0.9664	&	32.87	&	0.0584	&	0.8845	&	25.79	&	0.1951	&	0.6267	&	18.99	&	0.5600 &  3.47M   	\\
    \hline
     \multirow{6}{*}{MMVP}   &   $\times$  &   $\times$&  $ \times$&   $\checkmark$&   $\checkmark$	&0.9000	&	28.31	&	0.1874	&	0.9375	&	30.43	&	0.1342	&	0.8759	&	25.36	&	0.2304	&	0.6593	&	19.90	&	0.4992 &  2.75M \\
&   $\times$  &   $\times$&   $\checkmark$&  $ \checkmark$&  $ \checkmark$	&0.9284	&	30.14	&	0.1115	&	0.9667	&	32.79	&	0.0603	&	0.8937	&	26.11	&	0.1693	&	0.7159	&	20.71	&	0.3570 &  2.79M \\
&  $ \times $ &  $ \checkmark$&   $\checkmark$&   $\checkmark$&   $\checkmark$	&0.9296	&	30.22	&	0.1064	&	0.9669	&	32.87	&	0.0576	&	0.8965	&	26.26	&	0.1571	&	0.7199	&	20.76	&	0.3555 & 2.80M  \\
     \cline{2-18}
     
&   $\checkmark$  &   $\checkmark$&   $\checkmark$&   $\checkmark$&   $\checkmark$	&0.9296	&	30.29	&	\textbf{0.1051}	&	\textbf{0.9675}	&	32.99	&	\textbf{0.0567}	&	0.8958	&	26.22	&	\textbf{0.1554}	&	0.7175	&	20.76	&	0.3517 & 2.80M  \\
&   $\checkmark$  &  $ \times$&   $\checkmark$&  $ \checkmark$&   $\checkmark$	&\textbf{0.9300}	&	\textbf{30.35}	&	0.1062	&	0.9674	&	\textbf{33.05}	&	0.0580	&	\textbf{0.8970}	&	\textbf{26.29}	&	0.1569	&	\textbf{0.7203}	&	\textbf{20.84}	&	\textbf{0.3510} & 2.79M  \\

    \toprule
    \end{tabular}
    }
    
    \label{tab:ucf_result}
    \vspace{-2mm}
\end{table*}

Besides quantitative evaluation, we also showcase several qualitative visualizations in Figure~\ref{fig:qual} and compare them with SimVP~\cite{gao2022simvp} which achieved the second-best performance on our splits of the UCF Sports dataset. We especially select four samples in the hard and intermediate subsets for visualization. The first sample in Figure~\ref{fig:qual} shows a far-away object with long-distance movement. This is caused by the low fps of the dataset. MMVP accurately captures the displacement of the object while not losing the color and shape of the object. The challenge of the second sample in Figure~\ref{fig:qual} is also typical to UFC Sports, i.e. fast movement with motion blur. MMVP is able to recover the correct shape of the athletes' feet even if they are blurred in the observed frames. The third and fourth samples both show a fast camera movement, with and without large foreground movements. Camera moving with complex backgrounds is extremely difficult for video prediction. Even SimVP, which generated high-quality images for most of the datasets, barely captured the camera movements in these two samples and caused drastic blurs. Benefiting from its appearance-agnostic motion prediction module, MMVP achieved impressive performance in such cases.

\section{Discussion}
From Table~\ref{tab:ablation} and Table~\ref{tab:ucf_result} in Sec.~\ref{sec:exp}, we observe that the proposed MMVP framework can always achieve better or comparable performance with significantly fewer parameters compared to the other SOTA methods. To better understand the high efficiency of MMVP, we break down a video prediction system into three components: i) content encoder, which encodes the image sequences; ii) prediction-related modules, which take charge of predicting features for future frames based on the output of the content encoder; iii) content decoder, which decodes the output of the predicted features output by the prediction-related modules. Then we examine the model size for each component in three video prediction systems: STIP~\cite{chang2022stip}, SimVP~\cite{gao2022simvp}, and the proposed MMVP (configurations follow the second row and the fourth row in Table \ref{tab:ablation}); see Table \ref{tab:param_compare}. 

In Table \ref{tab:param_compare}, prediction-related modules in STIP~\cite{chang2022stip} and SimVP~\cite{gao2022simvp} handle both motion and appearance features. However, in MMVP, prediction-related modules specifically handle only the motion features, leaving all appearance features to the content encoder and decoder. The most noticeable fact from Table \ref{tab:param_compare} is that in STIP~\cite{chang2022stip} and SimVP~\cite{gao2022simvp}, the prediction-related modules take majority of the model parameters (the ratios are $99.7\%$ and $99.4\%$ respectively). In contrast, motion-related modules in MMVP only take 14.5\% and 15.9\% of the parameters. This fact supports our argument in Sec.~\ref{sec:intro} that decoupling motion prediction and appearance maintenance effectively avoid the cumbersome structures of the prediction modules and largely improve the prediction efficiency. Another observation is that despite the small size, the prediction-related modules in MMVP can still support high-quality motion prediction. This validates the efficiency of the motion matrices when describing the motion information, which results in a lightweight design of the prediction module.

\begin{table}[tbh!]
\caption{Model size breakdown. The numerical values are the number of parameters taken by each component in the video prediction systems. }
\centering \small
    \resizebox{\linewidth}{!}{%
    
    \begin{tabular}{c|cccc}
    \toprule
    Method &    Content Encoder &   Prediction Modules   &   Content Decoder    &   Total    \\
    \hline
    \hline
    STIPHR~\cite{chang2022stip}   &   29.86K   &   17994.8K   &  29.54K   &   18054.2K  \\
    SimVP~\cite{gao2022simvp}  &   7.5k   &   3447.1K   &   11.7K  &   3466.3K \\
    MMVP-mini  &   369.4K   &  113.2K   &   228.3K   &   710.9K \\
    MMVP   &   1472.2K   &   404.9K   &   911.9K  &   2789.0K \\

    \toprule
    \end{tabular}
    
    \label{tab:param_compare}}
\end{table}
\section{Conclusion}
The proposed Motion-Matrix-based Video Prediction framework (MMVP) is an end-to-end trainable two-stream pipeline. MMVP uses motion matrices to represent appearance-agnostic motion patterns. As the sole input of the motion prediction module in MMVP, the motion matrix can i) describe the many-to-many relationships between feature patches without training for extra modules; ii) intuitively compose future features with multi-scale image features through matrix multiplication. It helps the motion prediction become more focused, and efficiently reduces the information loss in appearance. Extensive experiments demonstrate the superiority of MMVP compared to existing SOTA methods in both the model size and performance.

{\small
\bibliographystyle{ieee_fullname}
\bibliography{egbib}
}

\end{document}